\documentclass[lettersize,journal]{IEEEtran}
\usepackage{wrapfig}
\usepackage{booktabs} 
\usepackage{array}    
\usepackage{caption}  
\usepackage{amsmath,amsfonts}
\usepackage{algorithmic}
\usepackage{algorithm}
\usepackage{array}
\usepackage[caption=false,font=normalsize,labelfont=sf,textfont=sf]{subfig}
\usepackage{textcomp}
\usepackage{stfloats}
\usepackage{url}
\usepackage{verbatim}
\usepackage{graphicx}
\usepackage{cite}
\usepackage{multirow}
\usepackage{xcolor}
\begin{document}

\title{Toward Safety-First Human-Like Decision Making\\ for Autonomous Vehicles in Time-Varying Traffic Flow }

\author{
Xiao Wang,~\IEEEmembership{Senior Member,~IEEE, Junru Yu, Jun Huang, Qiong Wu, 
\\Ljubo Vacic,~\IEEEmembership Life Senior Member,~IEEE, Changyin Sun,~\IEEEmembership Senior Member,~IEEE
}
\thanks{This paper was supported by the National Natural Science Foundation of China under Grant 62173329 and the University Scientific Research Program of Anhui Province (2023AH020005).}
\thanks{Manuscript received XXX XX, 2025; revised XXX XX, 2025. Corresponding author:  cysun@ahu.edu.cn}}



\maketitle

\begin{abstract}
Despite the recent advancements in artificial intelligence technologies have shown great potential in improving transport efficiency and safety, autonomous vehicles(AVs) still face great challenge of driving in time-varying traffic flow, especially in dense and interactive situations. Meanwhile, human have free wills and usually do not make the same decisions even situate in the exactly same scenarios, leading to the data-driven methods suffer from poor migratability and high search cost problems, decreasing the efficiency and effectiveness of the behavior policy. In this research, we propose a safety-first human-like decision-making framework(SF-HLDM) for AVs to drive safely, comfortably, and social compatiblely in effiency. The framework integrates a hierarchical progressive framework, which combines a spatial-temporal attention (S-TA) mechanism for other road users' intention inference, a social compliance estimation module for behavior regulation, and a Deep Evolutionary Reinforcement Learning(DERL) model for expanding the search space efficiently and effectively to make avoidance of falling into the local optimal trap and reduce the risk of overfitting, thus make human-like decisions with interpretability and flexibility. The SF-HLDM framework enables autonomous driving AI agents dynamically adjusts decision parameters to maintain safety margins and adhering to contextually appropriate driving behaviors at the same time. Extensive experiments in CARLA validate the framework’s superior performance, which enlarges the minimum TWHs by 41.8\% to keep safer distance away from others, while improves the average velocity by 2.5\%, reduces the average acceleration and yaw rate by 23.5\% and 60.5\%. The results highlight the potential of SF-HLDM to bridge the gap between machine-driven precision and human-like flexibility in AV systems, paving the way for more interpretable and socially acceptable autonomous driving solutions.
\end{abstract}


\section{Introduction}
\IEEEPARstart{H}{uman} have a remarkable ability to make use of every second and every inch while driving in complex traffic environments. For example, in busy urban traffic, human drivers quickly interpret numerous cues, such as distances to surrounding vehicles, potential hazards on the road, and the intention of other road users. They then integrate these cues into informed decisions that balance safety, efficiency, comfort and even social compliance\cite{r1}. This aptitude emerges from fundamental cognitive processes combined with the cumulative experience drivers gain over time\cite{r2}. By continuously refining their internal understanding of the driving context, humans exhibit flexibility and resilience in ever-changing traffic conditions.

In contrast, AVs mostly process with predefined rules or principles, thus limited the ability to deal with unexpected and emergent situations. Although recent development of end-to-end methods facilitates AVs advanced capabilities to deal complex traffic scenarios with multi-source situated data comprehensive processing\cite{r3}, such methods process data in black box and lack of understanding of dynamic environment semantics, thus increase the incomprehensibility and unpredictability of vehicle behavior to other human road users. In efforts to emulate such competence, research on human-like decision making for has steadily progressed\cite{r4}\cite{r77}. Many existing methods rely on sophisticated learning techniques, including supervised and reinforcement learning (RL), to help AVs handle a range of driving scenarios\cite{r5}\cite{r6}\cite{r7}\cite{r78}. However, humans do not always strictly follow the traffic laws, and the traffic laws do not always specify traffic behavior. Besides, even if a driver drives into the same traffic situation twice, he may make different decisions, which will change the situation state and trig diversified evolutions. As traffic densities increase and participants interact with each other more frequently, these methods of learning and imitating at behavior level reveal their limitations in migratability across different contexts\cite{r7}\cite{r8}. Therefore, elements that influence the decision-making, and why and how they make an influence need to be further analyzed to design interpretable human-like decision-making models for human-vehicle mutual understanding. 

A key reason for human success in these settings is the capacity to synthesize multiple sources of information and weigh potential outcomes. Briefly speaking, human drivers navigate dense traffic by systematically seeking advantages and avoiding drawbacks in light of accumulated experience, they keep an eye on the surrounding traffic participants, predict their intention, and assess the driving risks of making different responses, while considering their right to use the road space\cite{r9}. This interplay of situational awareness, personal judgment, and adherence to traffic rules enable humans to drive in a manner that is both appropriately cautious and dynamically responsive to fluctuating conditions. At the same time, they also factor in social norms and personal mood, including acknowledgment of ROW and courtesy to vulnerable traffic participants, to ensure that individual decisions align with the shared interests of other road users. Through repeated interaction with the environment, humans further sharpen their decision-making processes, whether in expected or unexpected scenes.

\begin{figure*}[t]
    \centering
    \includegraphics[width=\textwidth]{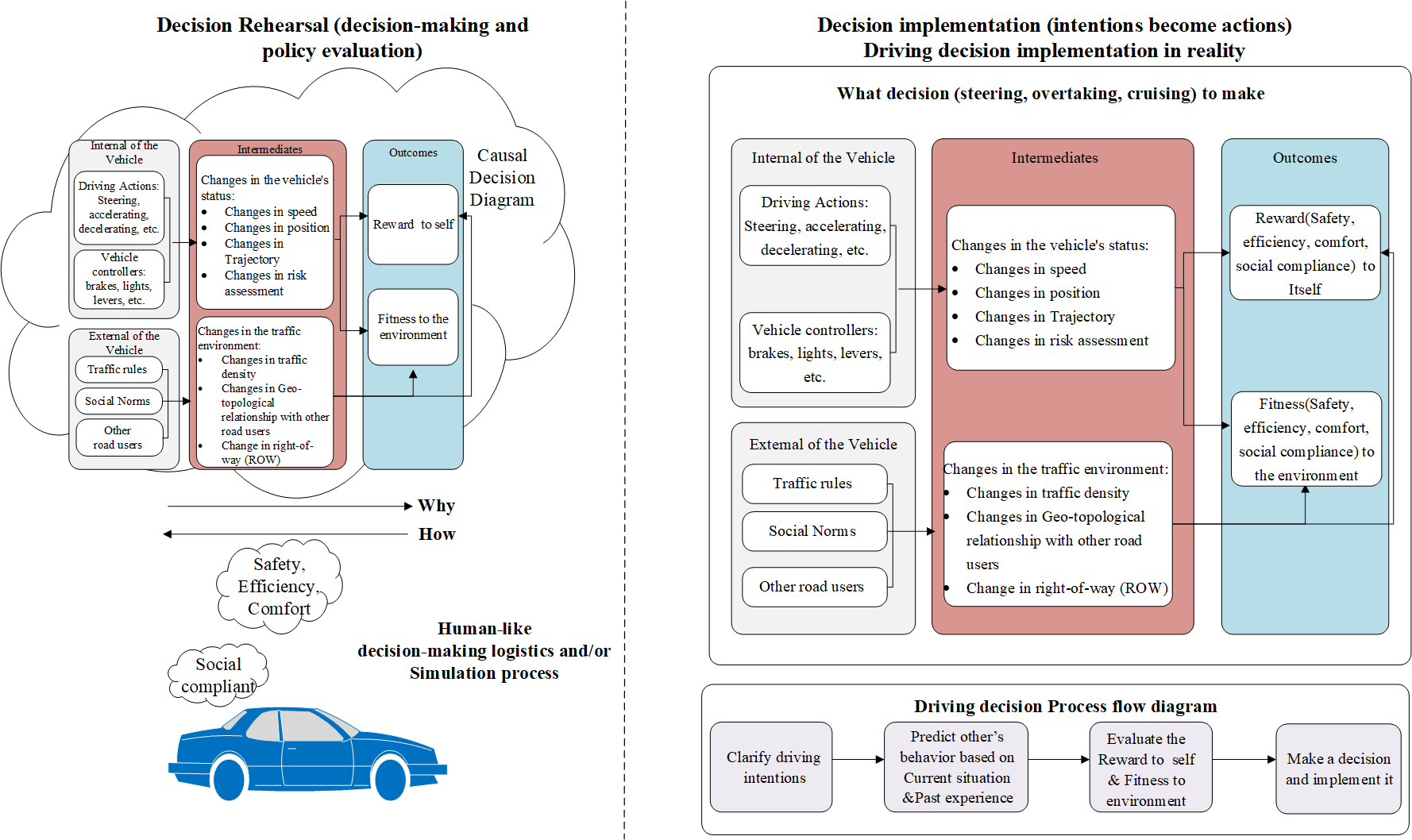}
    \caption{Analysis of human-like decision-making process of AVs}
    \label{fig:human-like}
\end{figure*}
In this paper, we propose a Safety-First Human-Like Decision Making framework for AVs drive safely, efficiently, comfortably with social compliance in time-varying traffic flow. To this end, We firstly analyze how human make a decision in traffic, as Fig.1 shows. The proposed framework is illustrated in Section III with detailed description. 

The novelty and significant contributions of this research includes:

\begin{enumerate}
    \item A multi-feature late fusion spatial-temporal mechanism is designed to capture other road users’ intention within the most appropriate time series, which impressively improves the intention inference accuracy and enhanced AV's situational awareness.
    \item The concept of absolute ROW area is proposed and described mathematically, based on which the ROW-violation index is defined, and AV's behavioral social compliance estimation is realized. Thus provide AVs with priori-knowledge for making decisions more socially compatible.
    \item The genetic algorithm is introduced into the decision-making module to optimize the weight matrix, thus expanding its search space more efficiently while make avoidance of falling into the local optimal trap.
\end{enumerate}

By considering the ROW belongingness when analyzing benefit and cost at each situated decision, the framework adapts decision parameters in real time to preserve safety margins while ensuring contextually appropriate driving maneuvers, thereby reflecting the way human drivers balance risk management, driving quality, and collective interests, to achieve reduced collision rates, improved efficiency and comfortableness, and smoother integration of AVs into modern traffic systems. 

\section{Research Foundations}

\subsection{Human Behavior in Traffic} 
Human drivers exhibit a nuanced decision-making process that combines driving experience, recognition of right-of-way (ROW), ongoing assessments of risks and rewards, while adherence to social conventions. This behavioral sequence can be understood through the lens of social behavioral theories, which posit that decisions are often shaped by a complex interplay of personal goals, perceived norms, and situational cues\cite{r10}. When seeking advantages and avoiding drawbacks, they take into account not only immediate conditions, such as vehicle spacing, potential hazards, and traffic signals, but also more implicit factors including intention of surrounding road users, social norms, and collective expectations\cite{r11}\cite{r12}\cite{r79}. This multifaceted process is remarkably adaptive and flexible, showing consistency across a wide range of traffic scenarios. Whether merging onto a highway, navigating a busy intersection, or responding to an unexpected pedestrian crossing, human drivers rely on these same foundational principles to prioritize safety while striving to reach their destinations efficiently. 

\subsubsection{Why Safety First}
Safety-first decision-making is naturally essential for human in traffic, given that they must operate in environments where the stakes of errors are extremely high, and the human body is very fragile to take any errors in traffic. Experts from industries have argued for “responsibility-first” approaches\cite{r13}, emphasizing strict adherence to legally assigned duties, such logic does not fully account for the fluid nature of real-world traffic. Generally speaking, humans do not strictly obey the traffic rules, for example, red-light-running\cite{r14} is a pretty normal but dangerous social behavior, but even so, AVs are supposed to give way to pedestrians in stead of hitting on them.

Conditions on the road are often uncertain and complex, with elements such as several kinds of surrounding traffic participants, full of different intention,  and the behavior of pedestrians and cyclists frequently fluctuating in ways that cannot be entirely captured by static rules. By placing safety first, AVs can proactively manage potential collisions and other undesirable events, thereby better serving the overarching objective of protecting human life. This risk-sensitive stance aligns with human instincts to prioritize harm avoidance when in doubt, a principle also supported by ethical and regulatory guidelines in automotive engineering\cite{r15}.

\subsubsection{What means to be Human-Like}
When we ask AVs to drive like a human, we are requesting it to possess a spectrum of cognitive and perceptual abilities that reflect the sensitivity, flexibility, adaptability and social awareness that human drivers naturally exhibit\cite{r16}. Such abilities include perceiving subtle context cues and adjusting behavior in real time, prioritizing safety while also maintaining an efficient flow of movement with social compatible behavior. They involve the capacity to gauge others' intention, interpret road users' implicit signals, negociate the right to use the road in specific time, and uphold common driving customs that extend beyond the written rules of the road. In order to equip AVs with these human-centric capabilities, we analyze how humans understand ROW attribution and violation while driving in fluctuating traffic surrounding by other road users, with consideration of social norms and traffic laws, which in turn enhances trust, predictability, and meaningful interpretability of automated actions for all who use or encounter these systems.

Driving like human offers multiple benefits for AVs. First, it improves predictability and social compatibility: other road users, no matter biological or mechanical, have come to expect certain patterns of behavior, such as subtle gestures of courtesy or slight speed adjustments to indicate cooperative intent. When AVs emulate these behaviors, their actions become more predictable and understandable to human drivers, cyclists, and pedestrians, thereby enhancing overall road safety and acceptance. Second, human-like driving accounts for the tacit knowledge that human drivers rely on knowledge of unspoken norms and situational nuances that cannot be entirely codified in traffic laws, such as always be courteous to vulnerable participants. Third, by adopting a decision-making strategy that is attentive to contextual factors such as evolving traffic dynamics, AVs can react more robustly to unexpected circumstances, leading to fewer edge cases and a smoother integration of autonomous systems into existing traffic networks. 

\subsection{The Spatial-Temporal Attention Mechanism}
Driving in time-varying traffic flow needs to pay attention to the geo-topological and social impact relationships between vehicles and key information within certain time series. Therefore, STA mechanism has recently gained prominence in the design of prediction modules for AVs\cite{r17}\cite{r18}\cite{r19}. Rooted in deep learning methods such as attention-based neural networks, S-TA enables the model to selectively focus on the most important inputs at both the spatial (e.g., lateral and longitudinal positions of vehicles or pedestrians) and temporal (e.g., recent movement trends, sudden changes in velocity) dimensions\cite{r20}\cite{r21}\cite{r75}. By integrating these forms of attention, the autonomous driving system refines its perception of the traffic environment, filtering out noise and prioritizing crucial cues for decision making\cite{r22}\cite{r23}. 

Recent works show that STA mechanisms significantly increase accuracy in predicting the behavior of surrounding traffic participants\cite{r24}\cite{r53}. In particular, attention-based architectures can highlight vehicles that pose an imminent risk, or emphasize periods of time when abrupt changes in velocity occur, thereby allowing the AV to plan safer and more efficient maneuvers\cite{r25}. This attention-driven trajectory computing framework of surrounding traffic participants complements traditional sensor fusion techniques and predictive models by giving the decision-making algorithm a dynamically focused view of the most critical aspects of the scene\cite{r80}. In dense traffic settings where interactions are multi-sourced, diversified and frequent, the STA approach helps identify potential ROW conflicts and predict the real-time intention of other drivers, thus reducing the likelihood of abrupt or dangerous maneuvers\cite{r28}\cite{r29}. ThAerefore, it provides AVs with the capability of making appropriate decisions by learning from better predictions.
\begin{figure*}[b]
  \centering
  \includegraphics[width=1.0\textwidth]{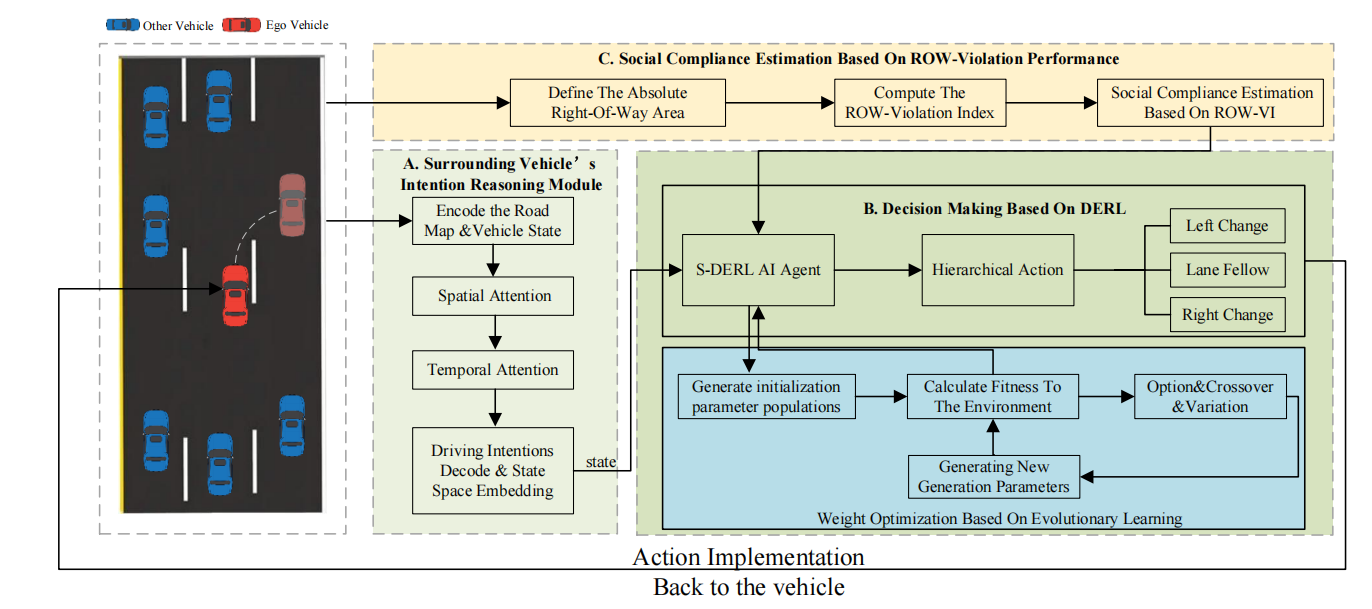}
  \caption{SF-HLDM based on Situation Awareness and Social Compliance}
  \label{fig:XXX}
\end{figure*}
\subsection{Deep Evolutionary Reinforcement Learning Methods}
Reinforcement learning (RL) offers a natural framework for enabling AVs to learn from continuous interaction with the environment, optimizing decision policies based on reward signals tied to safety, efficiency, and comfort\cite{r21}\cite{r81}\cite{r82}. State-of-the-art RL algorithms have shown promise in highway merging\cite{r22}\cite{r57}, lane changin\cite{r23}, and adaptive cruise control tasks\cite{r23}. They excel at balancing multiple objectives, such as minimizing travel time while respecting speed limits and maintaining safe following distances. However, these methods often face challenges in non-stationary and complex environments, which can cause them to converge to suboptimal or locally optimal strategies. Moreover, training efficiency becomes a bottleneck when dealing with high-dimensional state and action spaces, common in real-world driving scenarios, leading to significant computational overhead.

DERL combines the benefits of deep neural networks with evolutionary strategies to address the above mentioned limitations of RL. Whereas standard DRL may get stuck in local optima or require extensive hyperparameter tuning, the evolutionary component in DERL facilitates exploration over a broader range of policy parameters\cite{r56}. Specifically, evolutionary learning utilizes global search and evolutionary selection mechanisms to effectively mitigate overfitting risks while improving the model's generalization performance\cite{r25}\cite{r63}. Besides, it uses distributed asynchronous learning to leverage the scaling of computation and models that have been successful in other fields of AI. This augmented exploration can expand diversity of search space, and identify high-performing strategies that might be missed by gradient-based methods alone\cite{r26}\cite{r64}.

In the context of time-varying traffic flows, DERL enables the decision-making model to continuously adapt to changing conditions by choosing the policy with the highest fitness value\cite{r58}\cite{r70}. Precisely, by generating a population of candidate policies and evaluate their fitness to the environment, DERL incorporates novel actions that prove advantageous in complex or unforeseen scenarios, thus improving the overall robustness of the ego vehicle. Furthermore, DERL’s inherent parallelism and reduced sensitivity to hyperparameters can expedite training, making it more feasible to deploy learned policies in real-time traffic. Compared with spatial-temporal attention modules, DERL works more effectively and efficiently weigh the critical elements of driving scenarios, ultimately enhancing both safety and efficiency in decision making.

\subsection{Social Compliance Estimation}
Dense traffic frequently involves a wide array of road users, including cars, buses, motorcycles, bicycles, and pedestrians. Making a decision in such contexts should follow established legal and compliant frameworks while also accounting for situational nuances. These nuances may include whether a driver appears to cede space, how traffic participants coordinate road use with others at complex intersections, or whether pedestrians are on the curb waiting to cross\cite{r28}. Advanced decision-making models employ real-time environment perception and predictive analytics to update other participants' intention estimation continuously, ensuring that AVs can respond responsibly to dynamic changes in traffic flow.

Most current researches on social compliance in traffic assume human drivers exhibit varying degrees of altruism or egoism in traffic, and use the reward to self as reflection of diverse social-value orientations\cite{r29}\cite{r30}. More specifically, such method describes other road users as altruism or egoism based on data\cite{r65}, and to define the ethical performance of traffic participants\cite{r31}\cite{r60}. However, the driving behavior of each travel heavily influenced by the urgency of the task, thus we think it is not fare to judge them on the ethical level just based on data, especially in emergent situations. In addition to formal traffic laws, there are also implicit social norms to regulate human behavior when living in a society or driving in city. For example, while driving traffic, each participant has their own awareness of right-of-way (ROW)\cite{r66}. Drivers who consistently violate ROW by aggressively merging or cutting off others are commonly perceived as reckless or discourteous\cite{r68}. In social behavioral terms, these actions disrupt shared social norms and may provoke negative reactions from other drivers. While occasionally such maneuvers can reduce individual travel time, they increase the risk of accidents and broader inefficiencies due to sudden braking, lane changes, or evasive maneuvers by nearby vehicles. Therefore, the times and frequency of ROW-violation to others are more appropriate to use as driver social compliant performance evaluation. 

\section{METHOD}
In this SF-HLDM framework, we first design an intention inference model, define the absolute ROW area for each traffic participant and introduce the ROW-violation index to estimate the social compliance of the ego vehicle's driving behavior, and propose the social-compliant decision model based on deep evolutionary reinforcement learning method. The proposed hierachical progressional framework is illustrated in Fig.2, which consists of three parts: the intention inference module, social compliance estimation module and the evolutionary decision-making model. The intention inference module applies spatial-temporal attention mechanism to extract globle critical features from traffic scenarios in the recent continuous time sequences, encoding the spatial relationships among vehicles and the temporal evolution of their trajectories. And then decodes the driving intention of surrounding vehicles and embeds the inferred intention vectors into the state space of the decision-making layer. The decision-making model uses a social compliance and environment fitness based hybrid reward function to train each DRL agent, enhancing its performance and robustness in complex traffic environments. In which, the genetic algorithm is applied to optimize the S-DERL AI agent’s network weights, effectively mitigates overfitting risks while improving the model's generalization performance. This agent then operates with a hierarchical action space, where the high-level module generates abstract driving strategies (e.g., lane-keeping or lane-changing), and the low-level module outputs specific vehicle control commands (e.g., steering angle or acceleration).   

\subsection{Recognizing the Critical Participants based on Multi-Feature Late Fusion Special-Temporal Attention Mechanism} 
When human drivers are situated in a tricky environment, they typically make the next decision by considering a sequence of historical observations rather than just depend on the current observation, and extract the critical elements by assigning varying levels of importance to specific objects located at different postions at each time slide. Our framework incorporates a spatial-tmeporal multi-feature late fusion mechanism to imitate such recognition logics, shown in Figure 3. This mechanism contains two complementary modules: the spatial attention module and the temporal attention module. It dynamically assigns different importance value to different spatial regions and temporal segments, producing a comprehensive state vector that encodes vehicle interactions and trajectory changes. The state vector then serves as an essential input to the autonomous driving system’s decision-making layer. 

\subsubsection{The Spatial Attention Mechanism}

The spatial attention mechanism focuses on critical regions within an image, enabling the network to adaptively emphasize the most relevant spatial features for making decisions when driving in time-varying traffic flows. It operates between the convolutional layers and the recurrent layers.At each time step \emph{t}, the convolutional layers produce a set of L region vectors,$\{v_t^i \}_{i=1}^L$, where \emph{L=m×n} represents the total number of regions in the feature map, and each vector $ v_t^i \in \mathbb{R}^d $ encodes the features of a corresponding image region. These region vectors are then processed by the spatial attention mechanism to compute a context vector $z_t$, which captures the most relevant spatial features. The context vector $z_t$ is calculated as a weighted sum of all region vectors: 

\begin{equation}
z_t = \sum_{i=1}^L g_t^i \cdot v_t^i \label{eq:zt}
\end{equation}

where, $g_t^i$ represents the attention weight for the i-th region at time step t.

The attention weights are computed by an attention network $g_t^i$, which takes the region vector $v_t^i$ and the LSTM hidden state $h_t^i$ as input. The attention network is designed as a fully connected layer followed by a softmax function, ensuring that the weights are normalized across all regions:

\begin{equation}
g_t^i = \text{Softmax}(\omega_v \cdot v_t^i + \omega_h \cdot h_{t-1}) \label{eq:gt}
\end{equation}

where $\omega_v$ and $\omega_h$ are learnable parameters of the attention network.

The computed context vector$\ z_t$ serves as input to the LSTM layer in the global multi-feature late fusion mechanism. By reweighting the region features, the spatial attention mechanism acts as a dynamic mask over the CNN feature maps, selectively emphasizing the most informative regions for decision-making. This selective focus not only enhances the model's ability to interpret critical spatial features but also reduces the number of parameters, thereby improving the training and inference efficiency.

\subsubsection{The Temporal Attention Mechanism}

The temporal attention mechanism is employed to model the importance of information across different time steps, enabling the network to focus on the most relevant temporal segments. This mechanism operates over the outputs of the LSTM layer, which encode the temporal evolution of the driving scenario. For a sequence of LSTM outputs \(\{h_1, h_2, \dots, h_T\}\), the temporal attention mechanism assigns a scalar weight \(\omega_{T+1-i}\) to each output \( h_{T+1-i} \), where the weights are computed as:

\begin{equation}
\omega_{T+1-i} = \text{Softmax}(v_{T+1-i} \cdot h_{T+1-i}), \quad i=1,2,\dots,T \label{eq:wT}
\end{equation}

where, $v_{T+1-i}$ represents a feature vector learned during training, and the softmax function ensures that the weights are normalized across all time steps.

\begin{figure}[h]
    \centering
    \includegraphics[width=8.6cm]{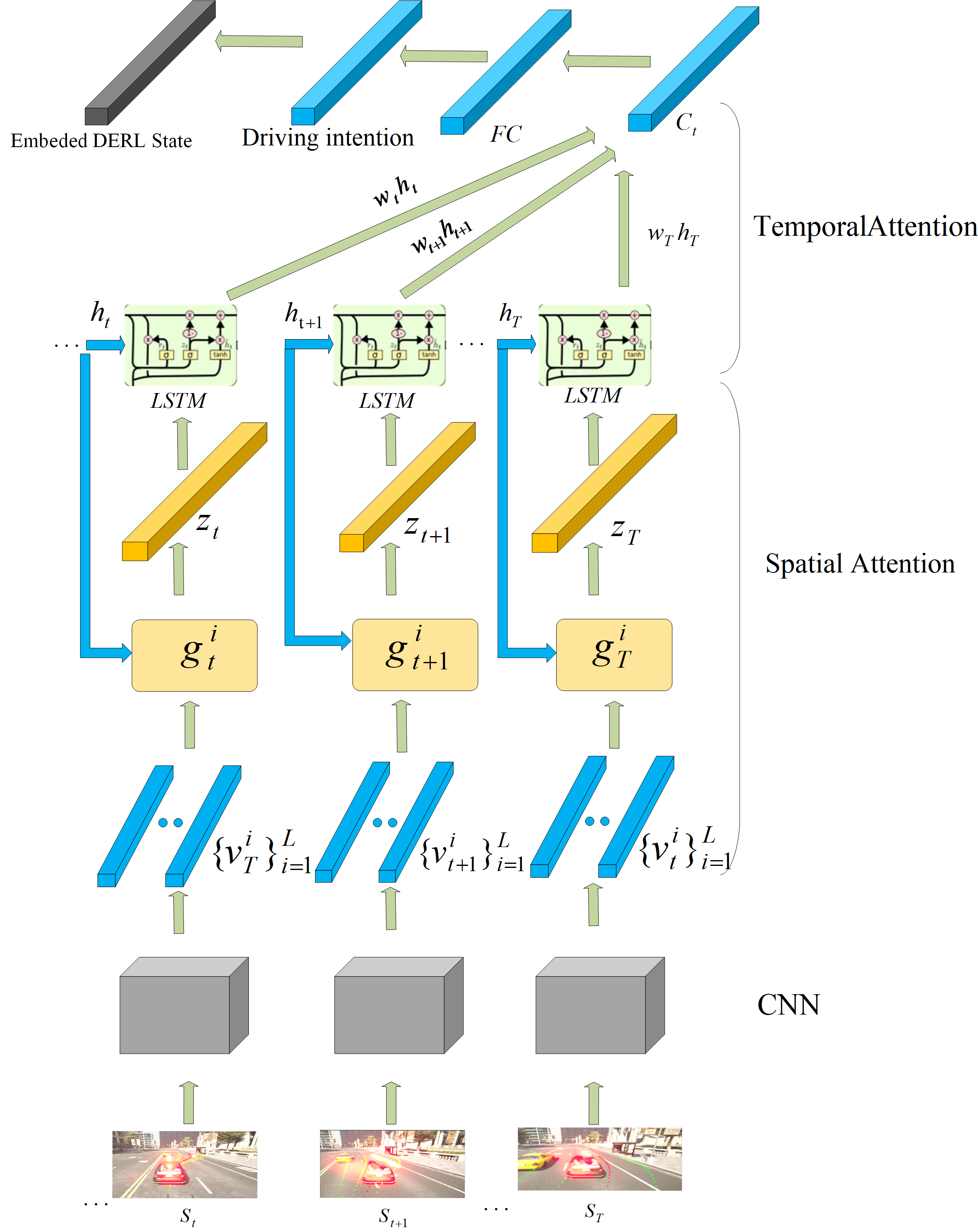}
    \caption{Spatial-Temporal Multi-Feature Late Fusion Mechanism}
    \label{}
\end{figure}

The context vector$\ C_T$, which summarizes the temporal information, is computed as a weighted sum of the LSTM outputs:

\begin{equation}
C_T = \sum_{i=1}^T (w_{T+1-i} \cdot h_{T+1-i}) \label{eq:CT}
\end{equation}

The context vector$\ C_T$ encapsulates the most critical temporal features by dynamically prioritizing outputs from important time steps, such as abrupt lane changes or sudden braking, thereby providing a refined representation of temporal information. A fully connected layer, FC, then further processes this vector, and the result is provided to the actor-network for hierarchical decision-making.

By including temporal attention, this mechanism adaptively concentrates on pivotal time steps and deemphasizes less informative periods. Such a design improves model robustness and interpretability, which are crucial in dynamic traffic environments where accurate temporal comprehension is essential for safe decision-making. 

Through the collaboration of spatial and temporal attention mechanisms, the framework dynamically models the significance of spatial features and temporal segments in complex traffic scenes, thereby enabling accurate driving intention inference. The spatial attention mechanism ensures flexible modeling of inter-vehicle interactions by emphasizing critical spatial features, while the temporal attention mechanism effectively captures the temporal evolution of vehicle behaviors, generating a holistic spatiotemporal representation.

This spatiotemporal attention mechanism significantly enhances the perception capabilities and robustness of autonomous driving systems in dynamic traffic scenarios, providing strong support for decision-making in complex environments.

\subsection{Deep Evolutionary Reinforcement Learning for Making the SCSE Decisions}

Deep reinforcement learning (DRL) has demonstrated considerable effectiveness in addressing complex sequential decision-making tasks, particularly in autonomous driving. In this research, we employ the Twin Delayed Deep Deterministic Policy Gradient (TD3) \cite{r32} algorithm as the foundation of our decision-making framework. The purpose of this framework is to make safe, comfortable, and socially compatible decisions in high efficiency, which we refer to as SCSE decisions. The TD3 algorithm functions as an AI Agent\cite{r83}, directly generating hierarchical driving action policies, such as left lane change, lane following, and right lane change. This effectively addresses the challenges of continuous action spaces in autonomous driving. These policies are derived from a structured process that integrates spatial\-temporal features and social compliance estimation to regulate the ego vehicle's behavior, making it perfectly fit the complex and time-varying driving environment.

To further improve the decision-making framework's performance, we implement a weight optimization mechanism based on evolutionary learning.The optimization process starts with generating an initial parameter population, followed by evaluating fitness in the driving environment. New generations of parameters are iteratively generated through selection, crossover, and mutation operations, effectively enlarge the search space, while decrease the model's overfitting risks. It ensures the decision-making model continually adapts to ever-changing complex traffic situations. Combining TD3 with evolutionary learning allows our framework to run like a driving AI Agent with the ability to balance safety, comfort, social compliance across various autonomous driving tasks in an efficient way.

The primary advantages of TD3 stem from its dual Q-network architecture and delayed target updates. The dual Q-networks independently estimate the action-value function, and then the lower one is choosed to reduce over-optimistic value predictions. Moreover, TD3 delays the updates to the target policy network, thereby promoting stability and robustness throughout the learning process. These features make TD3 a perfect choice for addressing the complexities inherent in autonomous driving, particularly within dynamic and uncertain traffic environments\cite{r34}.

\subsubsection{The S-DERL AI Agent's State Space}

In our network, the state space is designed to capture the critical elements in each complex traffic scene by encoding spatial and temporal features into a structured representation. Driving intention inference is integrated into the state space to provide a deeper understanding of the surrounding environment and the interactions between vehicles. A detailed description of the state space, including its features and components, is provided in Table I.

\begin{table}[h!]
\small
\centering
\caption{ State-Space Description}
\label{tab:state_space}
\resizebox{0.5\textwidth}{!}{%
\begin{tabular}{cc}
\toprule
\textbf{Parameter} & \textbf{Description} \\ 
\midrule
$x_0, y_0$ & The horizontal and vertical positions of the ego vehicle \\ 
$v_x, v_y$ & The lateral and longitudinal speed of the ego vehicle \\ 
$a_{x0}, a_{y0}$ & The lateral and longitudinal acceleration of the ego vehicle \\ 
$x_i, y_i$ & The horizontal and vertical positions of the surrounding vehicle \\ 
$v_{xi}, v_{yi}$ & The lateral and longitudinal speed of the surrounding vehicle \\ 
$a_{xi}, a_{yi}$ & The lateral and longitudinal acceleration of the surrounding vehicle \\ 
$SI_i$ & Assumption: intention of the surrounding vehicles \\ 
\bottomrule
\end{tabular}%
}
\end{table}

\subsubsection{The S-DERL AI Agent's Action Space}

To address the diverse requirements of autonomous driving, we employ a hierarchical action space that operates at both strategic and tactical levels. At the high level, the framework delineates three discrete driving decision policies that are mutually exclusive: left lane change, maintaining lane to follow the car ahead, and right lane change. At any given time step, the ego vehicle must select one of these high-level decision policies for execution.

Each selected high-level action requires the specification of two continuous-valued parameters to ensure the maneuver is performed safely and efficiently: heading angle and acceleration \ braking rate. 
The heading angle, constrained within a range of $[-0.5, 0.5]$, prevents large, potentially unsafe turning angles. Meanwhile, the acceleration/braking rate, ranging from $[-5, 5]$, modulates the vehicle’s speed, where positive values indicate acceleration, and negative values signify braking.

\subsubsection{Weight Optimization Based On Evolutionary Learning}

To enhance the decision-making capability of the above DRL framework, the genetic algorithm (GA) is introduced to optimize the weight parameters in the RL process. By simulating decision-evolution through operations such as selection, crossover, and mutation, GA carries out a global search of the parameter space, thereby reducing the tendency of DRL to converge to suboptimal solutions\cite{r36}. Notably, this approach does not rely on gradient information, facilitating efficient exploration of the complex parameter landscape. The fitness function is designed as a multi-objective evaluation metric, expressed as follows:

The fitness function is designed as a multi-objective evaluation metric, expressed as follows:

\begin{equation}
\begin{aligned}
\textit{Fitness} = & \, \omega_1 \cdot \textit{Safety} + \omega_2 \cdot \textit{Efficiency} \\
                   & + \omega_3 \cdot \textit{Comfort} + \omega_4 \cdot \textit{SCE\_Score}
\end{aligned}
\label{eq:fitness}
\end{equation}

where the four weight parameters \(w_1\), \(w_2\), \(w_3\), and \(w_4\) are normalized to ensure their sum equals 1, thereby preventing any single metric from disproportionately influencing the overall evaluation. Safety is quantified based on the frequency or severity of collisions, reflecting the vehicle’s ability to avoid accidents. Efficiency is measured by the average vehicle speed, capturing traffic throughput. Comfort is assessed via the average variation in acceleration, representing the smoothness of the driving experience. Finally, \text{SCE\_Score} evaluates the social compliance estimation (SCE) the ego vehicle, which is computed by ROW-violation times and severity during its trajectory. By adjusting these weight parameters, the fitness function can flexibly accommodate the requirements of different driving scenarios. 

The optimization process begins with the random initialization of a population, where each individual represents a parameter vector. Following this, the fitness function evaluates the performance of each individual in real-world traffic simulations. The selection operation uses a tournament strategy, where a subset of individuals is randomly chosen, and the one with the highest fitness is retained for the next generation. Crossover operations combine portions of the genetic information from two parents to produce offspring, thereby expanding the search space and improving population diversity. Mutation operations introduce random perturbations to a small number of genes, effectively preventing the algorithm from becoming trapped in local optima and enhancing exploration efficiency.

Through this iterative process, the population quality improves progressively, and GA converges towards a global optimum. The final optimized parameters are utilized to initialize the DRL module, providing a more favorable starting point for subsequent training. This hybrid approach effectively integrates the global optimization capability of GA with the adaptive learning characteristics of DRL, enabling the framework to better balance safety, efficiency, comfort, and ROW fairness in complex traffic scenarios. Ultimately, this strategy achieves robust and efficient decision-making, even in highly dynamic and uncertain environments that the ego vehicle has not meet before.

\subsection{Social Compliance Estimation Based On ROW-Violation Performance}
In this section, we demonstrate how the \text{SCE\_Score} is computed in detail. Firstly, the concept of absolute right-of-way area is proposed. This area represents the region where a vehicle has absolute priority access. Next, the ROW-violation index is calculated by analyzing vehicle occupacy behavior of other road users' \text{A\_ROW} area. Finally, social compliance is assessed based on the ROW-violation index, offering a quantitative measure of adherence to social norms.

\begin{figure}[h]
    \centering
    \includegraphics[width=\columnwidth]{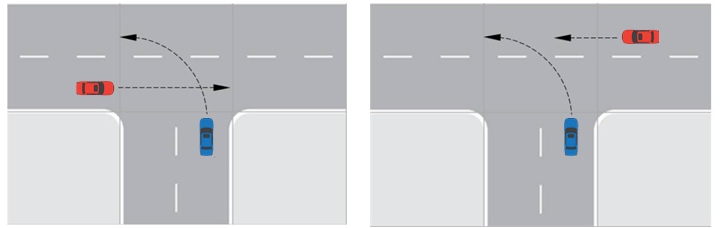}
    \caption{Illustration the Potential ROW Conflict Within B's Absolute ROW Area }
    \label{fig:reward_success_comparison}
\end{figure}

\subsubsection{Definition and Formalization of Absolute Right-of-Way}
To capture the essential constraints in complex vehicle interactions, we define the concept of \text{A\_ROW} as a spatially protected region surrounding a vehicle. As illustrated in Fig. 4, typical \text{A\_ROW} conflict scenarios—namely cross conflicts and merge conflicts—highlight the necessity of such a definition. Cross conflicts occur at the intersection center, where vehicle trajectories intersect at large angles, leading to severe traffic interference and high collision risk. Merge conflicts, by contrast, occur near the intersection exit, where vehicles from different entries aim for the same exit lane, resulting in less disruption but still frequent conflicts. 

The \text{A\_ROW} area provides a geometric boundary that other vehicles must not encroach upon, thereby preserving the right-of-way of the target vehicle. By explicitly modeling this constraint, the driving policy can make socially compliant and safety-aware decisions when faced with these typical conflict scenarios.These regions are mathematically defined as:
\begin{equation}
\begin{aligned}
\text{A\_ROW} &= \{(x, y) \mid x^{\min} \leq x \leq x^{\max}, \, y^{\min} \leq y \leq y^{\max}\} \\
\end{aligned}
\label{eq:Af_Ar}
\end{equation}

where:

\begin{equation}
\begin{aligned}
x^{\min} &= x, \quad x^{\max} = x + L \cdot \cos\theta, \\
y^{\min} &= y - \frac{\omega}{2}, \quad y^{\max} = y + \frac{\omega}{2}
\end{aligned}
\label{eq:coordinates}
\end{equation}

Here, \textit{x}, \textit{y} represent the vehicle’s position, \( \Theta \) is the heading angle, and \( \omega \) is the vehicle’s width. The stopping distance 
\( \mathcal{L} \) is defined as:

\begin{equation}
L = \frac{v^2}{2a_{\max}} \cdot \frac{1}{1 + k_{\rho} \cdot \rho}
\label{eq:length}
\end{equation}

where \( \nu \) is the velocity, \( a_{\max} \) is the maximum deceleration, 
\( \rho \) represents the road vehicle density, and \( k_{\rho} \) is a density adjustment coefficient.

The text{A\_ROW} dynamically updates based on the ego vehicle's motion state (e.g., velocity and heading angle) and the surrounding context (e.g., traffic flow). This ensures that the protected regions accurately reflect the vehicle’s immediate spatial constraints and the surrounding travelable regions.

\subsubsection{ROW Embedded Social Compliance Estimation}
Social compliance estimation plays a critical part in the text{Fitness} function. Usually speaking, the more social compatible of human driving behavior, the more safety to self and efficiency to the collective. 
To enhance the driving efficiency, safety, comfort, and compliance of AV operation beyond that of manual driving, the lane-changing model outlined in our paper adopts a multifaceted reward function. This function meticulously integrates several key performance indicators to guide the vehicle’s decision-making processes\cite{r48}. The reward function in this paper considers the following aspects:

Firstly, the model prioritizes speed performance by ensuring that the vehicle attains or closely approximates the reference speed \( \nu _{\text{ref}} \) within a permissible timeframe, as delineated in Equation (9). The speed performance metric is quantified by assessing the deviation between the actual vehicle speed and
\( \nu _{\text{ref}} \).

\begin{equation}
r_v = -\omega_v \cdot \frac{\sum_{t=0}^{T-1} t \lvert v_t - v_{\text{ref}} \rvert}{\sum_{t=0}^{T-1} t}
\label{eq:rv}
\end{equation}

Secondly, the model places a significant emphasis on comfort, evaluated through the analysis of acceleration variations between consecutive time intervals, following Equation (7).

\begin{equation}
r_c = -\omega_c \cdot \frac{\lvert a_t - a_{t-1} \rvert}{\text{time}_t - \text{time}_{t-1}}
\label{eq:rc}
\end{equation}

Regarding safety, the model incorporates the widely recognized Time-To-Collision (TTC) metric as a key safety criterion. The correlation between TTC values and the reward mechanism is inversely proportional: when TTC falls below 3.5 seconds\cite{r50}, lower values yield reduced rewards. This relationship highlights the importance of maintaining a safe following distance to minimize collision risks.

\begin{equation}
r_s = -\omega_s \cdot \max\{0, \frac{3.5 - t_{\text{ttc}}}{3.5}\}
\label{eq:rs}
\end{equation}

Regarding compliance, the model incorporates the {A\_ROW} criterion as a core measure. The reward mechanism is inversely proportional to {A\_ROW} violations: greater encroachments into AR-protected regions result in reduced rewards. This emphasizes adherence to spatial constraints and respect for other road users' safety zones.

To handle dynamic scenarios where {A\_ROW} regions may change (e.g., due to sudden braking), a Dynamic Right-of-Way Adjustment mechanism is introduced. The associated reward is defined as:

\begin{equation}
r_d = -\omega_d \cdot A\_ROW_{ij} \cdot \left(1 + e^{-\beta T}\right)
\label{eq:rd}
\end{equation}

where \( A\_ROW_{\text{ij}} \) denotes the change in the overlapping area between the ego vehicle and another vehicle's AR region, and \( \omega _{\text{d}} \) is the response coefficient. The \( e^{ \beta T }\) is time decay term, \(\beta \) determines the decay rate, \(\mathbb{T} \) is the time elapsed since the intersection state was detected. This component incentivizes the agent to respond promptly to intersection events. The penalty for significant changes in the overlapping area encourages swift and adaptive behaviors, such as deceleration or lane changes, to mitigate potential conflicts effectively.	

The Stop Signal defines a termination state triggered by events like collisions, departure from the roadway, or reaching the maximum simulation duration without incidents. Encounters with the termination state due to adverse events incur significant penalties, while remaining within safe operational boundaries throughout the simulation period is rewarded\cite{r51}. This approach not only penalizes risky behaviors but also reinforces sustained safe driving practices, aligning with objectives to enhance travel efficiency, safety, and comfort in AV operation.

\begin{equation}
r_t =
\begin{cases} 
100, & \text{safe\_arrived} \\
-60, & \text{collision} \\
-40, & \text{wrong lane}
\end{cases}
\label{eq:rt}
\end{equation}

The Social Compliance Estimatio(SCE) value is calculated as the weighted sum of Equations (9–13), representing the comprehensive evaluation of an autonomous vehicle’s performance in complex, dynamic environments. The reward function integrates five key components: speed performance, comfort, safety, rule compliance, and terminal state outcomes. Each component quantifies the quality of vehicle behavior from distinct perspectives. For instance, speed performance evaluates the vehicle’s efficiency in approaching the reference speed, comfort assesses the smoothness of acceleration transitions, safety measures the maintenance of a safe following distance, and rule compliance examines the vehicle’s ability to adapt to dynamic right-of-way allocations. Meanwhile, the terminal state reinforces safe driving behaviors by rewarding or penalizing outcomes such as safe arrivals or collisions.

The physical significance of the SCE lies in its ability to comprehensively quantify and integrate multiple performance indicators, providing a holistic assessment of driving efficiency, safety, and social compliance. By adjusting the weights of each component, the reward function achieves multi-objective optimization, ensuring sensitivity to behavioral deviations. This approach guides autonomous vehicles in making decisions that effectively balance efficiency, safety, and comfort in real-world scenarios.

\section{EXPERIMENTS}

In this section, we evaluate the performance of the proposed hierachical progressional framework in typical urban road scenarios of different traffic densities, verificating its ability to reduce collisions and increase driving speed with comfortness by making the fittest decisions in diverse time-varying traffic situations.

\subsection{Experimet Settings}

The simulation platform CARLA is used for model training and peformance evaluation. CARLA offers significant flexibility to modify the vehicle and traffic environment parameters, thereby enriches the verification cases and enables more efficient development without incurring real-world safety hazards. The datasets used in this study were generated within the CARLA urban traffic simulation environment, where each vehicle’s target lane and route were randomly assigned to approximate real-world conditions. Data sampling occurred at a frequency of 10 Hz, with the total simulation time set to 10,000 seconds to ensure sufficient historical trajectory information. The features employed in the driving intent recognition model include lane identification, longitudinal and lateral positions, and longitudinal and lateral velocities. These features were obtained via the CARLA API. For real-world implementation, similar data can be acquired through an AV’s perception module, supplemented by high-definition maps. 

Considering that humans make decisions based not only on current situations but also on what happend in the last few seconds, which enable them combine the critical information more comprehensively in developmental perspectives, we think it is also inspiring for AVs to make better decisions. But, no one has ever tested that how long should we take into consideration, not mention to how long should we anble AVs to look-back. To determine the optimal length of historical trajectory time window for model training, various experiments were conducted to test and evaluate their impact on the performance metrics, including precision, recall, and F1-score. The results show in Fig.5 indicate that a 4-second time window enables the STA module achieve the best performance across all metrics(depicted in Fig. 5), providing the most effective balance between information sufficiency and noise reduction. 

\begin{figure}[h]
    \centering
    \includegraphics[width=8.6cm]{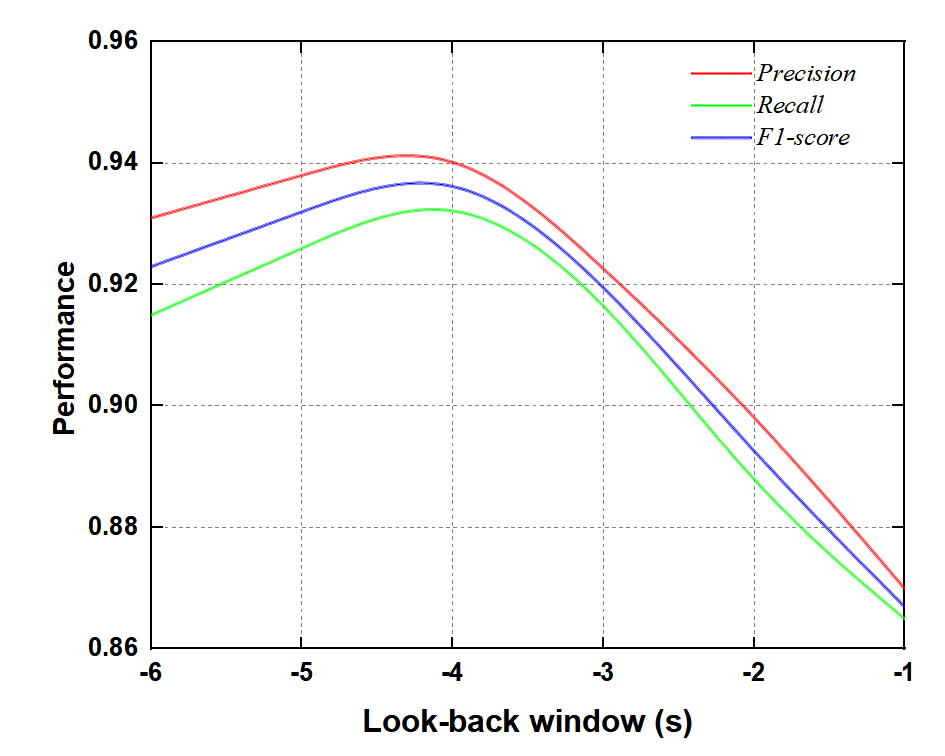}
    \caption{Performance metrics on different Look-Back time window lengths}
    \label{}
\end{figure}

Therefore, a 4-second historical trajectory window was used for the STA model training, and a down-sampling strategy was applied to achieve balance among left lane change, lane-keeping, and right lane change actions. The dataset was split into training and testing sets at an 8:2 ratio to facilitate model validation and performance assessment.

\begin{figure}[h]
    \centering
    \includegraphics[width=8.6cm]{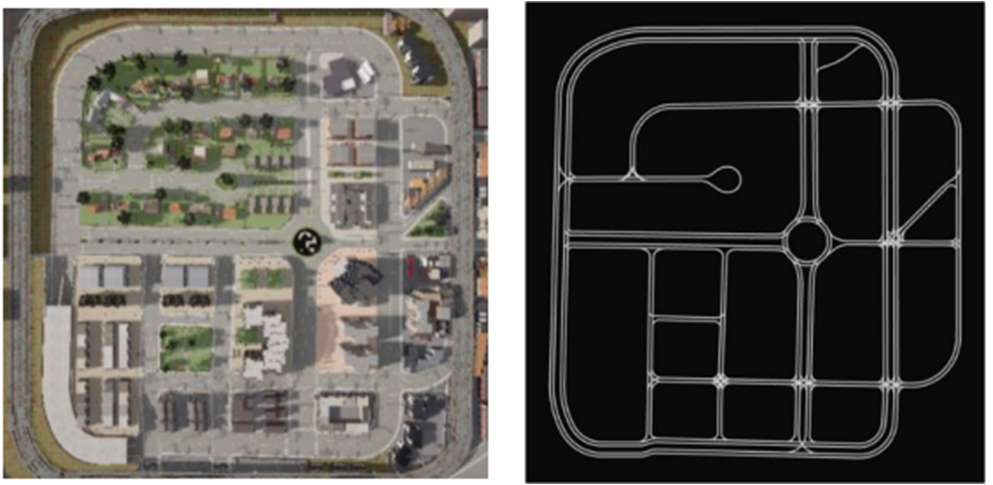}
    \caption{Bird’s eye view of Town3 in CARLA}
    \label{}
\end{figure}

We train all models in CARLA’s Town 3 map to ensure a fair comparison and evaluate the effectiveness of our approach and all baseline models. As shown in Fig. 6, this town features a highly dynamic and intricate urban layout, designed to mimic real-world suburban and urban driving scenarios. It encompasses several interconnected regions, including a suburban residential area, a central business district with multilane roads, and a series of complex roundabouts and intersections. The intricate network of Town 3 makes it particularly challenging for human-like flexible and adaptive decision-making, as it offers a wide range of driving conditions, such as long straight roads, sharp turns, elevated sections, and a variety of lane configurations and traffic flow patterns. These characteristics provide an ideal environment for assessing both fundamental driving skills and sophisticated decision-making capabilities.

To create a more realistic and challenging environment, we populate the town with 100 vehicles running in autopilot mode, randomly spawned across the map and managed using CARLA’s built-in traffic manager. These vehicles follow traffic rules, respond to traffic lights, and perform basic collision avoidance, creating a dynamic traffic flow that significantly increases the complexity of the learning task for our RL agent. The agent must now handle various interactive scenarios such as car following, overtaking, and yielding to other vehicles. Additionally, the ego vehicle is randomly positioned in a lane with an initial speed that avoids collisions at the outset, and its desired speed is set within the range of 10\text{--}18 \,\text{m/s}, with each simulation step corresponding to 0.1 seconds of real time. This setup not only mirrors real-world urban driving conditions but also challenges the RL agent to develop more robust and adaptive driving strategies.

All simulation and training processes were deployed across four Nvidia RTX 4090 GPUs, fully utilizing their parallel computing capabilities. Table II summarizes the key hyperparameters employed in the experiments.

\begin{table}[h!]
\centering
\caption{Evolutionary Algorithm and Deep Reinforcement Learning Hyperparameters}
\label{tab:hyperparameters}
\resizebox{\columnwidth}{!}{%
\begin{tabular}{cc}
\toprule
\textbf{Item} & \textbf{Value} \\ 
\midrule
Population Size & 50 \\
Selection Method & Tournament Selection \\
Tournament Scale & 5 \\
Selection Pressure & 0.8 \\
Crossover Probability & 0.8 \\
Crossover Method & Single-Point Crossover \\
Mutation Probability & 0.05 \\
Mutation Magnitude & $N(0, -0.1)$ \\
Maximum Generations & 100 \\
Single-objective Optimization Weights & 1 \\
Replay Buffer Size & $10^6$ \\
Discount Factor & 0.97 \\
Delayed Update Rate & $10^{-3}$ \\
Max Training Steps & 20,000 \\
Actor and Critic: Number of Features in the Hidden State & 128 \\
LSTM: Number of Features in the Hidden State & 64 \\
Learning Rate Schedule & Anneal Linearly to $10^{6}$ \\
\bottomrule
\end{tabular}%
}
\end{table}

\subsection{Driving Intention Inference Performance of the Multi-Feature Late Fusion STA Model}

We compare our proposed STA mechanism with other five Driving Intention Inference baselines, which are:

HMM-BF\cite{r40}: a driving intent inference method based on Hidden Markov Models. It takes behavioral feature sequences of vehicle motion and environmental context as input to infer the most likely driving intention.

HMM-AIO\cite{r5}: an extension of HMM-BF that integrates all-in-one feature representation, including behavioral features, traffic context, and driver-specific data, to enhance intention inference in complex scenarios.

LSTM-RNN\cite{r42}: which uses Long Short-Term Memory networks to process sequential data, such as vehicle motion and surrounding traffic states, capturing long-term dependencies to output driving intention.

Bi-LSTM\cite{r43}: an extended version of LSTM that processes sequences bidirectionally, considering both historical and future information, making it suitable for complex scenarios requiring bidirectional context.

SlowFast\cite{r44}: it is adapted from the SlowFast network for video analysis, the "Slow" pathway captures long-term patterns, while the "Fast" pathway focuses on short-term dynamics. Combined multi-scale information is used to infer driving intent.

While recent advances in transformer-based models have demonstrated promising performance in various sequential modeling tasks, the baseline methods selected in this study—including HMM variants and LSTM-based architectures—are representative and widely adopted in the field of driving intention inference. These models provide strong interpretability, established performance benchmarks, and lower computational overhead, making them suitable for both research comparisons and real-world deployment. Additionally, transformer-based methods typically require large-scale labeled datasets and high computational resources, which may not be practical in real-time autonomous driving scenarios or under limited data conditions. Our goal is to evaluate the effectiveness of the proposed STA mechanism against classical and widely-accepted baselines. Incorporating transformer-based approaches will be considered as part of future work as the domain continues to evolve.

The comparisve results of our STA mechanism (illustrated in Fig. 3) with the above mentioned five models. As can be seen from Table III, in the driving intention inference task, our proposed STA mechanism outperforms other methods on all comparative metrics, with improvements of 3.4\% on precision, 4.8\% on Recall-Rate, 4.1\% on F1-Score, and 2.3\% on accuracy.

\begin{table}[h!]
\centering
\caption{Results of Driving Intention Recognition}
\label{tab:driving_intention}
\resizebox{\columnwidth}{!}{%
\begin{tabular}{ccccc}
\toprule
\textbf{Model} & \textbf{Pr (\%)} & \textbf{Re (\%)} & \textbf{F1 (\%)} & \textbf{Acc (\%)} \\ 
\midrule
HMM-BF    & $77.2 \pm 1.7$ & $75.2 \pm 1.6$ & $76.2 \pm 1.6$ & $79.3 \pm 1.5$ \\
HMM-AIO   & $79.4 \pm 2.3$ & $76.8 \pm 1.3$ & $78.1 \pm 1.7$ & $80.1 \pm 2.2$ \\
LSTM-RNN  & $82.0 \pm 1.5$ & $80.5 \pm 1.0$ & $81.2 \pm 1.2$ & $84.5 \pm 1.3$ \\
Bi-LSTM   & $86.2 \pm 1.0$ & $85.1 \pm 1.5$ & $85.6 \pm 1.2$ & $87.2 \pm 0.9$ \\
SlowFast  & $89.5 \pm 1.0$ & $87.9 \pm 1.3$ & $88.7 \pm 1.1$ & $91.5 \pm 1.2$ \\
\textbf{Our proposed}      & \textbf{$92.5 \pm 2.0$} & \textbf{$92.1 \pm 1.5$} & \textbf{$92.3 \pm 0.8$} & \textbf{$93.6 \pm 0.8$} \\ 
\bottomrule
\end{tabular}%
}
\end{table}

Fig. 7 presents the normalized confusion matrix, providing a detailed assessment of the proposed STA mechanism in inferring L Trun, R Trun, and Straight driving intention. The diagonal elements represent the per-intention accuracy, demonstrating strong performance with 94\% for L Trun, 95\% for R Trun, and 92\% for Straight. The off-diagonal elements illustrate misclassification rates. Notably, 6\% of L Trun intention are incorrectly classified as R Trun, and 8\% of Straight intention are misclassified as R Trun. It suggests a potential for improvement in distinguishing R Trun from other intention, which deserves further exploration in the future work. This high overall accuracy demonstrates the effectiveness of the proposed method for driving intention inference.

\begin{figure}[h]
    \centering
    \includegraphics[width=8.6cm]{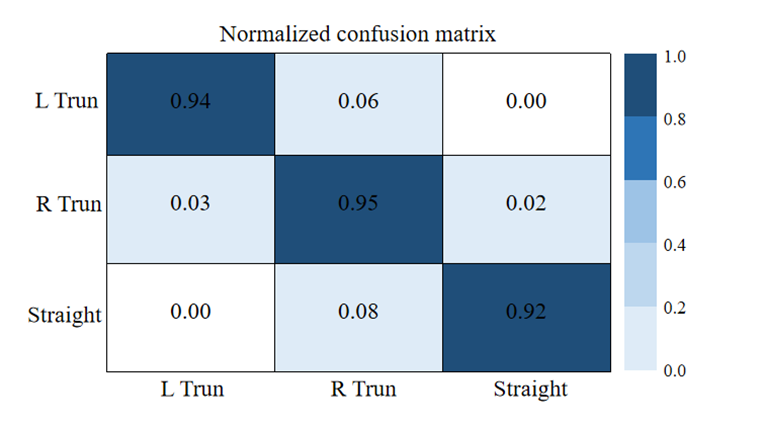}
    \caption{Normalized Confusion Matrix of the Driving Intention Inference Model based on Spatial-Temporal Attention}
    \label{}
\end{figure}

\begin{figure}[h]
    \centering
    \includegraphics[width=\columnwidth]{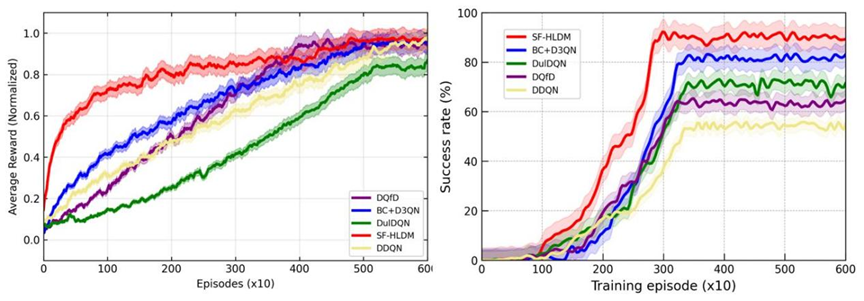}
    \caption{Comparison of Average Reward and Success Rate Across Training Episodes of Different Algorithms}
    \label{fig:reward_success_comparison}
\end{figure}
\subsection{Safer Decision-Making with Superior Evolutionay Performance }

This section demonstrate how SF-HLDM perform by comparing with other state-of-the-art decision-making baseline models:

DDQN\cite{r45}: double Deep Q-Network (DDQN) reduces Q-value overestimation by decoupling action selection and Q-value evaluation. It employs two separate networks, one for selecting the action and another for evaluating its Q-value, resulting in more stable and accurate decision-making.

DQfD\cite{r46}: deep Q-learning from Demonstrations (DQfD) incorporates expert demonstration data into the reinforcement learning process. By pre-training with expert trajectories and combining supervised learning with Q-learning, it accelerates convergence and enhances performance in complex tasks.

BC+D3QN\cite{r47}: behavioral Cloning with Double Dueling DQN (BC+D3QN) integrates behavioral cloning for pre-training and Double Dueling DQN for fine-tuning. The pre-trained policy guides the learning process, while D3QN improves decision accuracy by focusing on advantage value learning.

DulDQN\cite{r48}: Dual Deep Q-Network (DulDQN) extends Double DQN by enhancing the learning of advantage values, enabling it to achieve better stability and robustness in challenging environments with complex decision requirements.

All the models were trained under identical parameter configurations to facilitate fair comparisons, and the training performance curves are depicted in Fig 8. Each experiment was conducted over 100 episodes to assess their performance. To benchmark the models, a series of commonly used vehicle trajectory features were analyzed, including average speed, minimum time headway (THW), average acceleration, and average yaw rate. To ensure uniformity across all experiments, the same random seed was employed. Multiple rounds of experiments were then carried out on the trained models to evaluate their consistency, with the outcomes summarized in Table IV.

The model’s performance are further evaluted based on six metrics: average velocity, average acceleration, average yaw rate, minimum time headway (THW), average number of ROW-violation, and average number of lane changes. To maintain consistent conditions, a single random seed was applied across all experiments, and the results are presented in Table IV and Figure 8. The proposed SF-HLDM model consistently outperforms the above baseline models on every metric, achieving the highest average speed (\(15.440 \, \text{m/s}\)), with the lowest average accelertion (\(0.395\, \text{m/s\textsuperscript{2}}\)), the lowest yaw rate (\(0.015 \, \text{rad/s}\)), the lowest average number of lane changes (\(1.1\)) and the highest minimum THW (\(1.133\,\text{s}\)). Compared to other SOTA methods, the SF-HLDM framework improves the average velocity by 2.5\%, reduces the average acceleration and yaw rate by 23.5\% and 60.5\%, while enlarge the minimum TWHs by 41.8\%. 

\begin{table*}[ht]
\centering
\caption{Performance Metrics of Different Models}
\renewcommand{\arraystretch}{1.5} 
\setlength{\tabcolsep}{6pt}       
\begin{tabular}{lcccccc}
\toprule
\textbf{Model} & \textbf{\parbox{2cm}{\centering Average\\ velocity (m/s)}} & \textbf{\parbox{2cm}{\centering Average \\ acceleration (m/s\textsuperscript{2})}} & \textbf{\parbox{2cm}{\centering Average \\ yaw rate (rad/s)}} & \textbf{\parbox{2cm}{\centering Minimun \\ TWHs (s)}} & \textbf{\parbox{2.5cm}{\centering Average Number of \\ ROW violations}} & \textbf{\parbox{2.5cm}{\centering Average Number of \\ lane changes}} \\
\midrule
DDQN    & 12.707 & 0.516 & 0.132 & 0.388 & 8.8 & 5.5 \\
DQfD    & 12.844 & 0.634 & 0.053 & 0.520 & 8.2 & 5.7 \\
BC+D3QN & 15.059 & 0.682 & 0.038 & 0.799 & 5.6 & 2.8 \\
DulDQN  & 11.704 & 2.154 & 0.145 & 0.280 & 9.8 & 4.3 \\
SF-HLDM & \textbf{15.440} & \textbf{0.395} & \textbf{0.015} & \textbf{1.133} & \textbf{1.8} & \textbf{1.1} \\
\bottomrule
\end{tabular}
\label{tab:performance_metrics}
\end{table*}

It is worth noting that SF-HLDM enables the ego vehicle drive with faster average speed while keep longer distance away from other surrounding traffic participants, in impressive smoother and more gentle style. From a safety point of view, the larger the TWHs between vehicles, the safer it is. However, if the THW is too large, it will affect traffic flow and is not conducive to improving transportation efficiency. Existing researches on road user response timing via a large number of experiments indicate that the observed average response times varied widely between studies (0.5 s to 1.5 s) \cite{r54}. Therefore, considering AVs driving in mixed traffic full of human-driven vehicles, driverless vehicles with different degrees of automation, and even vulnerable traffic partipants, it is of great importance to have larger TWHs to keep safety-first. The experimental results clearly articulate that SF-HLDM generate the largest TWH while equip the AV with the fastest average velocity, lowest average acceleration and yaw rate. 

In addition, the average number of ROW-violations and average number of lane changes are captured. Lane-change is one of the most important reason for causing traffic accidents \cite{r55}, and decreasement in average number of lane changes will greatly lower the possibility of traffic accidents, which will be studied in our future researches. Meanwhile, when someone intent to make a lane change, the driver's behavior may accompanied by violating others' ROW, thus increase the possibility of social conflicts. The social compliance performance of each model can be estimated by these two indexes, which explicitly shows that the SF-HLDM framework runs with much more social compatible driving behavior. Statistically speaking, the SF-HLDM framework generates 1.8 action policies violating other participants' ROW, while other methods generate at least 5.6 action policies, improves the social compliance behavior by 67.9\%. Therefore, the SF-HLDM effectively balances the ego vehicle's driving quality and the overall traffic flow, highlighting its adaptability to fluctuating traffic conditions and underscoring its overall advantages for safe, efficient, smooth and social compliant driving behavior.

\subsection{Decision-Making Performance in Different Traffic Densities}

To assess the performance of the proposed model under varying traffic flow densities, a total of 100 driving experiments were conducted at three distinct traffic density levels: low density (\(60 \, \text{vehicles/km}\)), medium density (\(100 \, \text{vehicles/km}\)), high density (\(150 \, \text{vehicles/km}\)). The corresponding results are presented in Table V.

Table V summarizes the experimental results comparing the proposed SF-HLDM algorithm with the baseline BC+D3QN across low, medium, and high traffic density scenarios. The results demonstrate that SF-HLDM consistently outperforms the baseline across various key metrics. Under all traffic densities, SF-HLDM achieves significantly higher minimum Time to Worst-Case Hazard (TWHs), showcasing its enhanced capacity to predict and address potential risks. Additionally, the algorithm maintains a higher average velocity, reflecting its more efficient decision-making process and superior traffic flow management. SF-HLDM also registers lower average yaw rates, which contributes to greater stability and smoother vehicle handling, especially in high-density traffic. The results highlight SF-HLDM's strong adaptability and reliability in dynamic, complex traffic environments, underscoring its potential for deployment in real-world autonomous driving systems. Furthermore, the algorithm's robustness across different traffic conditions makes it a promising candidate for improving the safety and efficiency of autonomous vehicle operations.

Figure 10 presents selected keyframes from the CARLA simulation, showcasing the agent's decision-making process during continuous lane-change and overtaking maneuvers.The first lane-change decision takes place in round time of 4 s.The EV initially occupies the rightmost lane with its velocity
accelerating from zero to its desired level of 13 m/s. Then the
EV faces the slower LV and gradually decreases its velocity as
capped by the velocity of the LV. The EV equipped with our
proposed planner starts a lane change to the adjacent left lane,
which is the closest available lane considering the right boundary
of the current lane. This lane-change decision enables the EV
to raise its velocity up to 16.6 m/s which is designated by the
velocity limit. At around the time of 7 s, the EV encounters an
LV whose velocity is at 6 m/s. The decrease in the EV’s velocity
is observed due to the encountering of the slow-moving LV. In the
meanwhile, there are NVs on the left lane. Considering that the
NV is traveling at a lower velocity than the EV, the EV executes
the lane-change decision by the planner and brings the velocity
back to about 13 m/s. Subsequently, the EV finds itself trailing
again behind a slow-moving LV and the decision process is more
complex. The EV is given the options including lane keeping,
left lane change, and right lane change. In addition to the LV on
the same lane, there is an LV on the right lane and an NV on
the left lane. The EV is provided with a decision to change to
the left lane by the proposed planner at around the time of 10 s.
Finally, at around the time of 13 s, the EV chooses to go to the
left lane in a situation where the current lane and the adjacent
right lane are both occupied. At the end of the simulation, the
EV manages to navigate through this dynamic scenario occupied
with multiple SVs from the right-most lane to the left-most lane
where no SV is present. During the whole simulation, the EV
adjusts its velocity smoothly. The EV is able to maintain a high
traveling efficiency, and resume its velocity after the slow-downs
due to the encountering of the slow LVs and the lane-change
maneuvers.

The SF-HLDM framework’s ability to dynamically adjust parameters in real-time allows it to maintain safe margins while adhering to socially acceptable behaviors. By leveraging deep evolutionary reinforcement learning, the model avoids local optima, effectively balances competing objectives such as safety, comfort, and efficiency, and ensures robust performance in high-density traffic. This capability enables the model to replicate human-like decision-making processes, fostering smoother integration of autonomous vehicles into real-world traffic systems.

\begin{table*}[ht]
\centering
\caption{Performance Metrics of Different Models under Different Vehicle Densities}
\renewcommand{\arraystretch}{1.5} 
\setlength{\tabcolsep}{4pt}
\resizebox{\textwidth}{!}{ 
\begin{tabular}{lccccccc}
\toprule
\textbf{Vehicle Density} & \textbf{Model} & \textbf{\parbox{2cm}{\centering Average\\ velocity (m/s)}} & \textbf{\parbox{2cm}{\centering Average \\ acceleration (m/s\textsuperscript{2})}} & \textbf{\parbox{2cm}{\centering Average \\ yaw rate (rad/s)}} & \textbf{\parbox{2cm}{\centering Minimum \\ TWHs (s)}} & \textbf{\parbox{2.5cm}{\centering Average Number of \\ ROW violations}} & \textbf{\parbox{2.5cm}{\centering Average Number of \\ lane changes}} \\
\midrule
\multirow{5}{*}{Low Density}  
& DDQN & 13.739 & 0.509 & 0.119 & 0.489 & 8.4 & 5.1 \\
& DQfD & 12.997 & 0.588 & 0.049 & 0.591 & 8.1 & 5.2 \\
& BC+D3QN & 15.154 & 0.679 & 0.040 & 0.810 & 5.2 & 2.8 \\
& DulDQN & 12.334 & 1.776 & 0.121 & 0.357 & 8.9 & 4.5 \\
& SF-HLDM & \textbf{16.840} & \textbf{0.394} & \textbf{0.015} & \textbf{1.135} & \textbf{1.4} & \textbf{1.1} \\
\midrule
\multirow{5}{*}{Medium Density}  
& DDQN & 12.699 & 0.515 & 0.131 & 0.377 & 8.9 & 5.3 \\
& DQfD & 12.841 & 0.641 & 0.055 & 0.534 & 8.1 & 5.8 \\
& BC+D3QN & 13.993 & 0.793 & 0.054 & 0.703 & 6.6 & 2.6 \\
& DulDQN & 11.672 & 2.066 & 0.132 & 0.291 & 9.7 & 4.0 \\
& SF-HLDM & \textbf{14.876} & \textbf{0.461} & \textbf{0.027} & \textbf{0.991} & \textbf{1.9} & \textbf{1.0} \\
\midrule
\multirow{5}{*}{High Density}  
& DDQN & 10.533 & 0.881 & 0.166 & 0.329 & 9.5 & 6.2 \\
& DQfD & 10.587 & 0.897 & 0.083 & 0.443 & 9.3 & 6.0 \\
& BC+D3QN & 12.066 & 0.980 & 0.104 & 0.517 & 8.1 & 2.5 \\
& DulDQN & 9.002 & 2.432 & 0.174 & 0.231 & 10.8 & 5.8 \\
& SF-HLDM & \textbf{13.449} & \textbf{0.557} & \textbf{0.081} & \textbf{0.813} & \textbf{2.2} & \textbf{0.9} \\
\bottomrule
\end{tabular}
}
\label{tab:performance_metrics}
\end{table*}

\begin{figure}[h]
    \centering
    \includegraphics[width=\columnwidth]{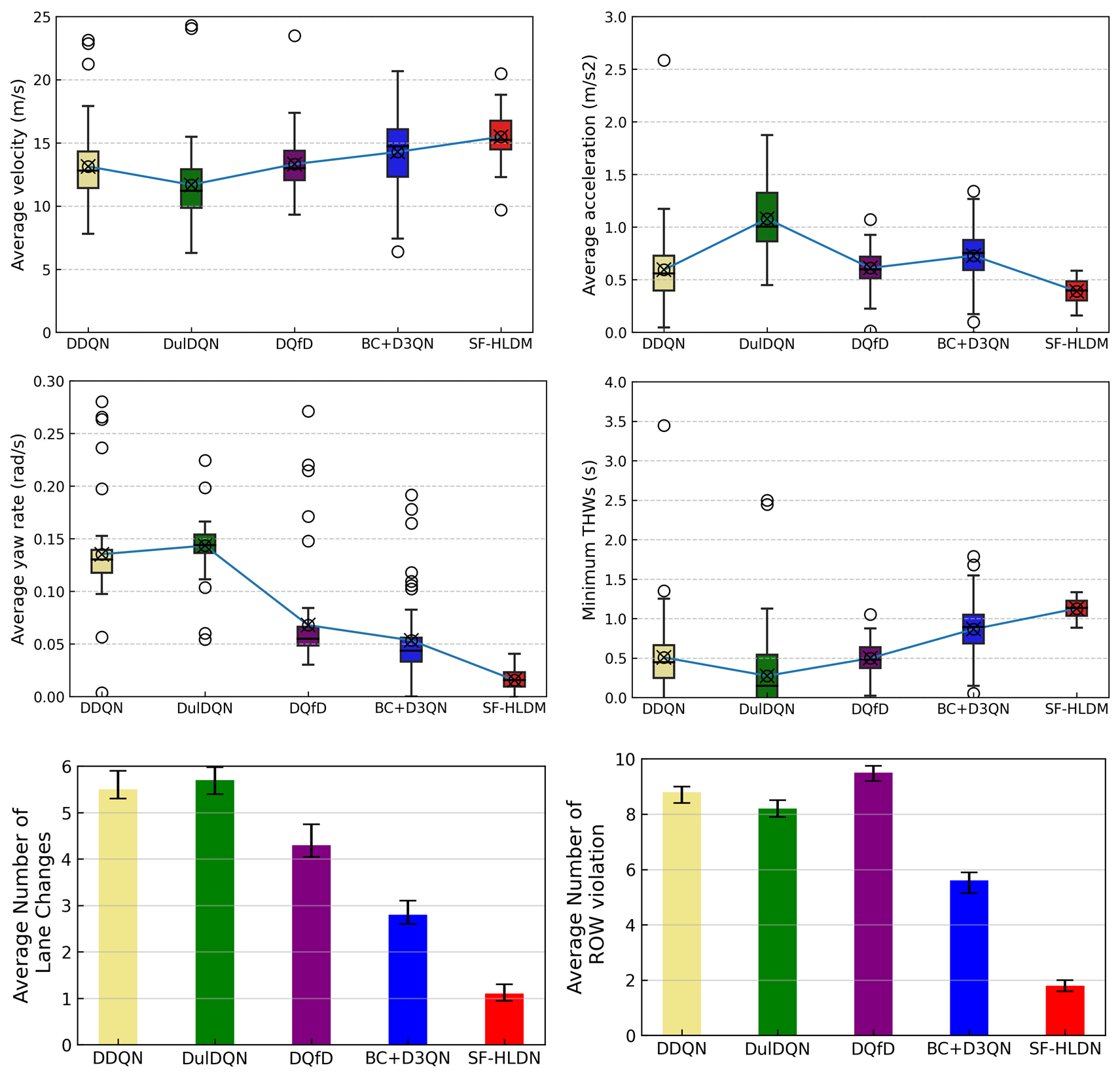}
    \caption{Comprehensive evaluation results of vehicle trajectories in 100 driving experiments}
    \label{fig:vehicle_trajectories}
\end{figure}

\begin{figure*}[b]
  \centering
  \includegraphics[width=1.0\textwidth]{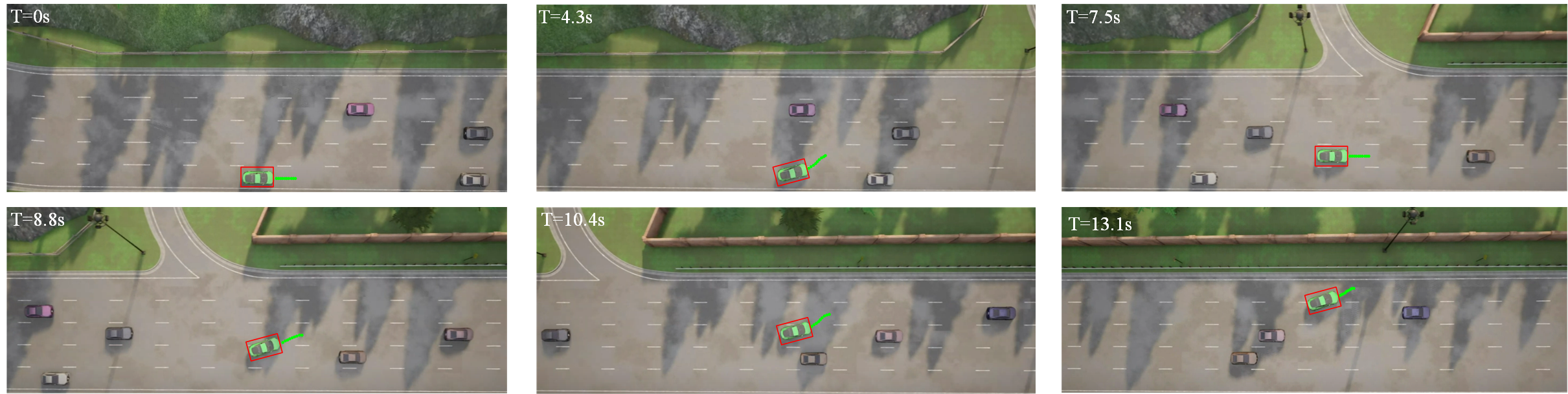}
  \caption{SF-HLDM based on Situation Awareness and Social Compliance}
  \label{fig:XXX}
\end{figure*}

\subsection{Effect of Critical Modules on SF-HLDM Performance}
To evaluate the contributions of the Situation Awareness Module and the Evolutionary Learning-Based Weight Optimization Module in the SF-HLDM model, we conducted an ablation study using the same experimental setup described in this Section . Specifically, the scenario initializes 100 NPC (Non-Player Character) vehicles to create a dynamic and time-varying traffic flow, ensuring realistic testing conditions.

In the ablation experiments, SF-HLDM-S represents the model with the Situation Awareness Module removed, which excludes spatial attention, temporal attention, and driving intention inference, allowing us to assess the impact of situation awareness on the decision-making process. SF-HLDM-E, on the other hand, removes the Evolutionary Learning-Based Weight Optimization Module and uses fixed or non-adaptive weights, enabling an evaluation of the role of evolutionary optimization in enhancing model performance. As shown in Fig. 11 and Table VI.

\begin{figure}[htbp] 
\centering
\includegraphics[width=0.5\textwidth]{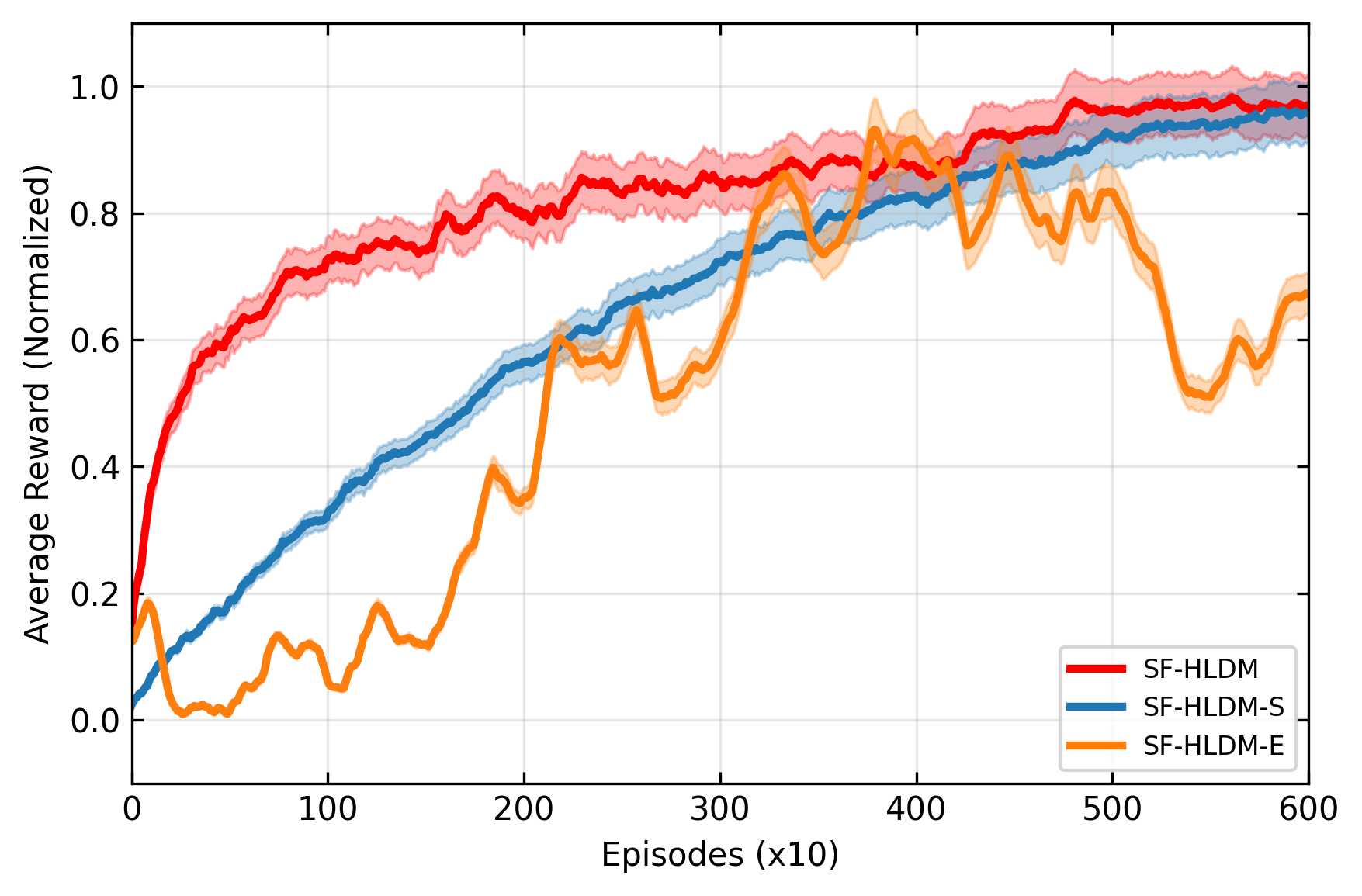} 
\caption{Multi-frame tracking of vehicles in dynamically time-varying traffic environments}
\label{fig:car_tracking}
\end{figure}

\begin{table*}[ht]
\centering
\caption{Performance Metrics of Different Models}
\renewcommand{\arraystretch}{1.5} 
\setlength{\tabcolsep}{6pt}       
\begin{tabular}{lcccccc}
\toprule
\textbf{Model} & \textbf{\parbox{2cm}{\centering Average \\ velocity (m/s)}} & \textbf{\parbox{2cm}{\centering Average \\ acceleration (m/s\textsuperscript{2})}} & \textbf{\parbox{2cm}{\centering Average \\ yaw rate (rad/s)}} & \textbf{\parbox{2cm}{\centering Minimum \\ TWHs (s)}} & \textbf{\parbox{2.5cm}{\centering Average Number of \\ ROW violations}} & \textbf{\parbox{2.5cm}{\centering Average Number of \\ lane changes}} \\
\midrule
SF-HLDM-S & 14.866 & 0.421 & 0.022 & 1.002 & 2.2 & 1.3 \\
SF-HLDM-E & 13.799 & 0.458 & 0.045 & 0.899 & 2.5 & 1.4 \\
SF-HLDM   & \textbf{15.440} & \textbf{0.395} & \textbf{0.015} & \textbf{1.133} & \textbf{1.8} & \textbf{1.1} \\
\bottomrule
\end{tabular}
\label{tab:performance_metrics}
\end{table*}
Fig.11 displays the performance of three methods: SF-HLDM, SF-HLDM-S, and SF-HLDM-E, based on the normalized average reward over 6000 episodes (scaled by a factor of 10). The red curve (SF-HLDM) consistently outperforms the others, achieving rapid convergence and maintaining the highest reward levels throughout the training process. The blue curve (SF-HLDM-S) improves steadily but lags behind SF-HLDM, reflecting moderate convergence speed and lower final rewards. In contrast, the orange curve (SF-HLDM-E) fluctuates significantly and shows slower overall progress, indicating instability and delayed convergence. Shaded regions around each curve represent the standard deviation, emphasizing the robustness of SF-HLDM compared to the other methods. This analysis highlights SF-HLDM’s effectiveness in achieving higher and more stable rewards.

Table VI makes a further comparision of SF-HLDM-S, SF-HLDM-E and SF-HLDM, which clearly articulates SF-HLDM improves the average velocity by \( 3.9\% \), reduces the average acceleration by \( 6.2\% \), and significantly decreases the yaw rate by \( 31.8\% \). The minimum TWH is extended by \( 13.1\% \), demonstrating the critical role of situation awareness in maintaining safe distances and smooth control. Similarly, when compared to SF-HLDM-E, which removes the evolutionary weight optimization submodule, SF-HLDM exhibits a \( 11.9\% \) improvement in average velocity, a \( 13.8\% \) reduction in yaw rate, and a \( 26.0\% \) increase in minimum TWH, highlighting the effectiveness of adaptive optimization in enhancing driving performance.

In summary, the results demonstrate that the full SF-HLDM model delivers a well-balanced performance, achieving the highest velocity while ensuring safety and smoothness, thanks to the integration of situation awareness and evolutionary optimization modules. The ablation studies underline the importance of these components in shaping the model's overall effectiveness.

\section{Conclusion}
This study proposes the Safety-First Human-Like Decision-Making (SF-HLDM) framework for autonomous vehicles, addressing critical challenges in time-varying and interactive traffic environments. The SF-HLDM framework integrates spatial-temporal attention mechanism, right-of-way awareness, and deep evolutionary reinforcement learning to enable AVs to emulate human-like decision-making processes. By dynamically adjusting decision parameters, this model balances safety margins with contextually appropriate driving behaviors, ensuring adaptability and resilience in complex traffic conditions.

Extensive simulations demonstrate the efficacy of SF-HLDM in improving safety, reducing right-of-way violation, and enhancing overall driving quality with better social compliant behavior. Compared to existing models, SF-HLDM offers superior interpretability and flexibility, bridging the gap between human-like adaptability and machine-driven precision. The findings underscore the importance of incorporating socially compatible decision-making elements, to address both situational awareness and environment fitness, into AV frameworks.

The proposed approach not only contributes to safer and more efficient autonomous driving but also enhances the integration of AVs into mixed traffic environments by fostering mutual understanding and trust between humans and machines. Future work will focus on expanding this framework to accommodate additional variables, such as cultural differences in traffic norms, and exploring real-world deployment scenarios to further validate its robustness and scalability. Through this effort, the SF-HLDM framework sets a promising path toward socially acceptable, safety-first autonomous driving systems.

\section{ACKNOWLEDGEMENT}
This work was supported by the National Natural Science Foundation of China under Grant 62173329 and the University Scientific Research Program of Anhui Province
(2023AH020005).

\bibliographystyle{IEEEtran}

\begin{IEEEbiography}[{\includegraphics[width=1in,height=1.25in,clip,keepaspectratio]{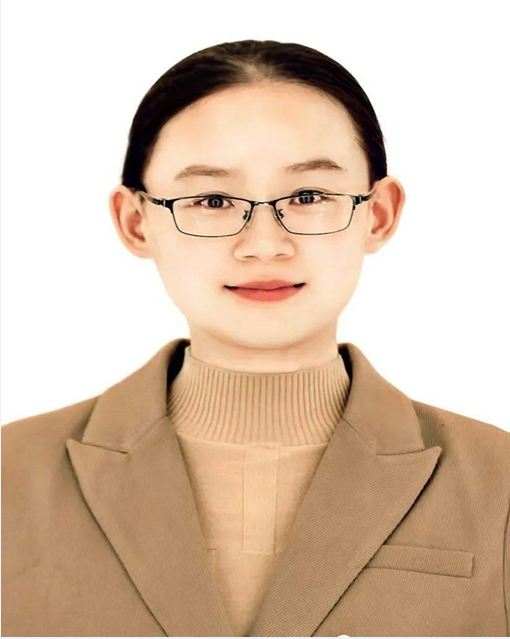}}]{Xiao Wang} (Senior Member, IEEE) received the Bachelor’s degree in network engineering from Dalian University of Technology, Dalian, China, in 2011, and the Ph.D. degree in social computing from the University of Chinese Academy of Sciences, Beijing, China, in 2016. She is currently a professor with the School of Artificial Intelligence, Anhui University, Hefei, and Vice President of the Embodied Intelligence Research Institute. Her research interests include social trans portation, parallel driving, cognitive and embodied intelligence, and multiagent modeling and collaboration. 
\end{IEEEbiography}
\vspace{10mm} 
\begin{IEEEbiography}[{\includegraphics[width=1in,height=1.25in,clip,keepaspectratio]{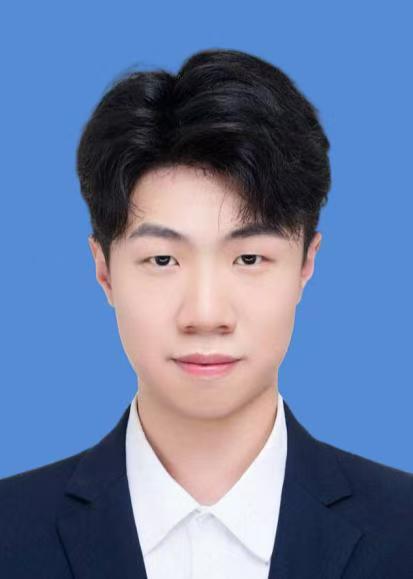}}]{Junru Yu} received the Bachelor of Engineering degree in 2021 and is currently pursuing a Master's degree at Anhui University in Hefei, Anhui, China. His primary research interests include autonomous decision-making methods for self-driving vehicles with adaptive traffic flow density.
\end{IEEEbiography}

\vspace{10mm}
\begin{IEEEbiography}[{\includegraphics[width=1in,height=1.25in,clip,keepaspectratio]{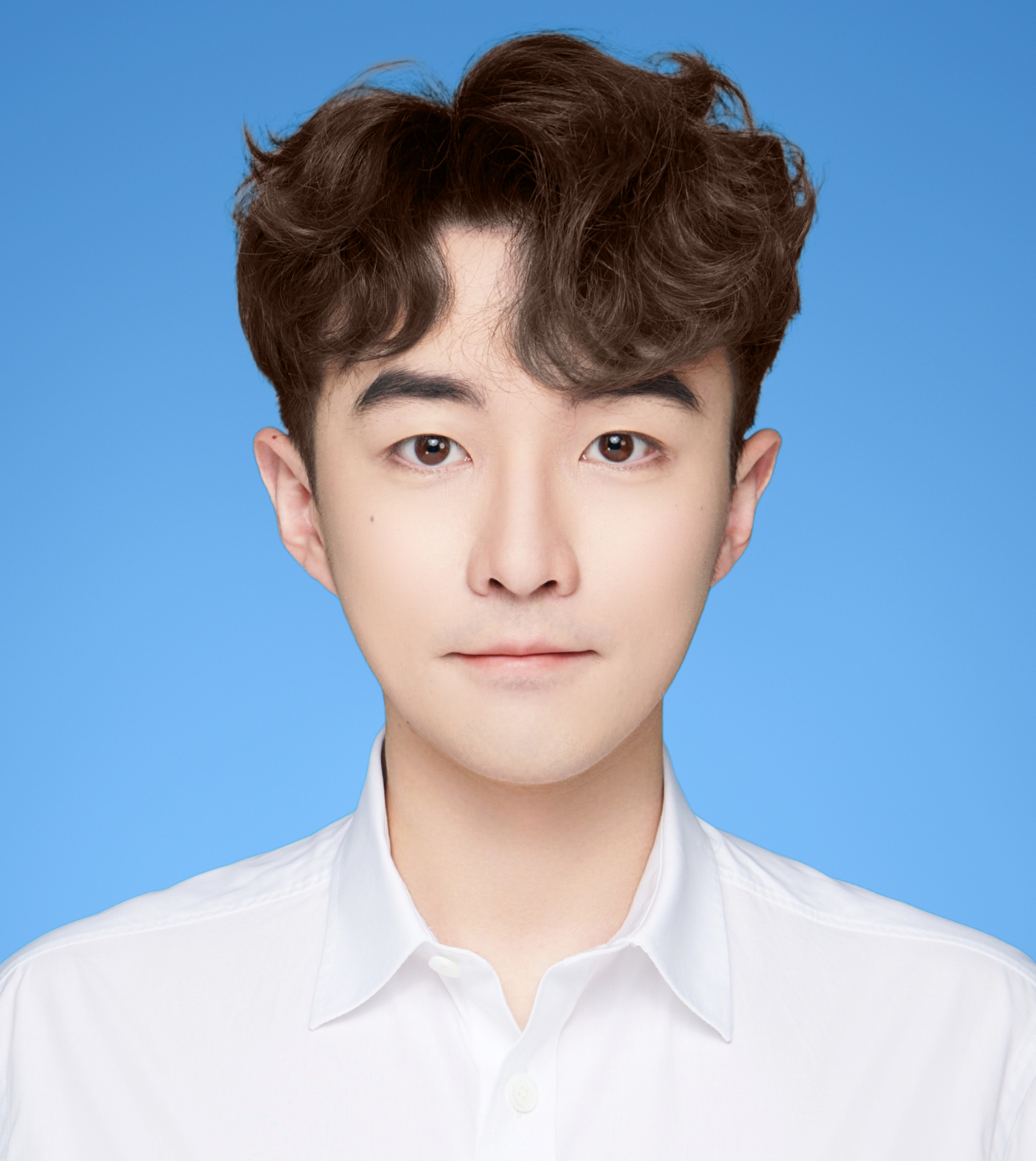}}]{Jun Huang} received his Master's degree from the Faculty of Science and Engineering, The University Of Manchester, Manchester, UK. He is currently pursuing the Ph.D. degree at the Department of Engineering Science, Faculty of Innovation Engineering, Macau University of Science and Technology, Macao, 999078, China. His research focuses on deep learning algorithms and reinforcement learning,  including autonomous vehicle trajectory prediction and planning, parallel Intelligence and prompt engineering.
\end{IEEEbiography}
\vspace{10mm}
\begin{IEEEbiography}[{\includegraphics[width=1in,height=1.25in,clip,keepaspectratio]{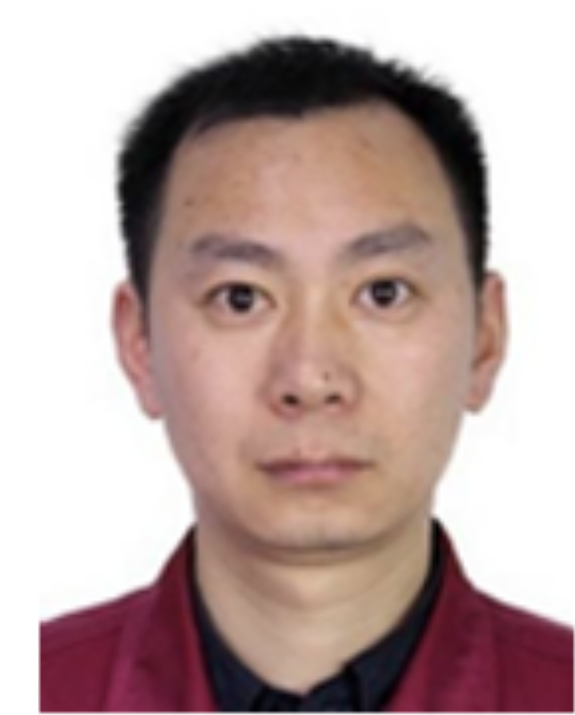}}]{Qiong Wu} received the Ph.D. degree in physicalscience from University of Science and Technologyof china,Hefei, Anhui,China,in 2008. He is aSenior Engineer with 10years of experience inintelligent driving system design and produet devel-opment, currently working with the JAC Group inHefei, China. He has undertaken 5 major science andtechnology projects, including Anhui Province's la-jor Science and Technology Projects and EmergingIndustry Special Projects. He has received two thirdprize for scientific papers and two third prize forScientifc and Technological Achievements in Anhui Province, and holds morethan 50 authorized invention patents. His research interests include intelligentvehicle control and brake-by-wire system.
\end{IEEEbiography}
\vspace{10mm}
\begin{IEEEbiography}[{\includegraphics[width=1in,height=1.25in,clip,keepaspectratio]{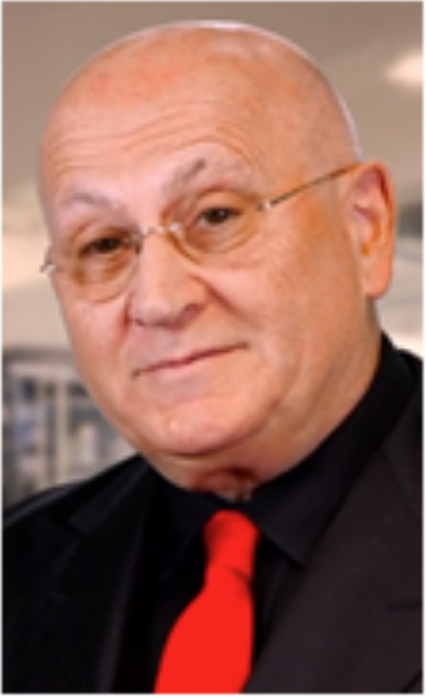}}]{Ljubo Vacic} (Life Senior Member, IEEE) receivedthe Graduate Diploma degree in electrical engineer-ing and the M,Phil. and Ph.D. degrees in controlsystems engineering from the University of Sarajevo,Sarajevo, Bosnia and Herzegovina, in 1973, 1976,and 1986, respectively. He is a control systems scien-tist and a practitioner, renowned for his contributionsto cooperative driverless vehicles and intelligent con-trol systems research and development. His researchachievements made news headlines and were broad-cast through media outlets throughout the world.Currently, he is: (i) President of the lEEE-Intelligent Transportation SystemsSociety; (ii)Distinguished Visiting Professor at the Southeast University,China;(iii)Director on the Board of Directors of the ANZCC SteeringCommittee;(iv)General Chair of the 2025 Australian and New ZealandControl Conference; (v) General Chair of the lEEE- Intelligent TransportationSystems Conference; and (vi)The Immediate Past Editor-in-Chief of IEEE-Intelligent Transportation Systems Magazine.
\end{IEEEbiography}
\vspace{-10mm} 
\begin{IEEEbiography}[{\includegraphics[width=1in,height=1.25in,clip,keepaspectratio]{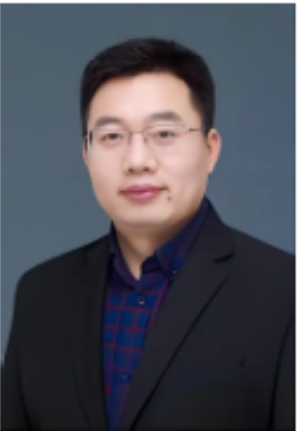}}]{Changyin Sun} (Senior Member, IEEE) received the bachelor's degree in applied mathematics from Sichuan University, Chengdu, China, in 1996, and the master's and ph.D. degrees in electrical engineering from Southeast University, Nanjing, Chian, President of Anhui University. He is the president with School of Artificial Intelligence, Anhui University. And he is also the president of Anhui University. His research interests include intelligence control, flight control, pattern recognition, and optimal theory.
\end{IEEEbiography}

\newpage

\end{document}